\documentclass[10pt,conference,a4paper]{IEEEtran}
\usepackage[utf8]{inputenc}
\usepackage{caption}
\usepackage{amsmath,graphicx}
\usepackage{subcaption}
\usepackage{amsfonts}
\usepackage{comment}
\usepackage{algorithm,algpseudocode}
\usepackage{todonotes}
\usepackage{geometry} 
\newgeometry{left=13.1 mm,right=13.1 mm,top=19.1 mm,bottom=39.9 mm}

\IEEEoverridecommandlockouts 
\begin{document}
\title{Variational Information Bottleneck Model for Accurate Indoor Position Recognition}


\author{
    \IEEEauthorblockN{Weizhu Qian\IEEEauthorrefmark{1}, Franck Gechter\IEEEauthorrefmark{1} \IEEEauthorrefmark{2}}
    \IEEEauthorblockA{\IEEEauthorrefmark{1}CIAD UMR 7533, Université Bourgogne-Franche-Comté, UTBM, 90010, Belfort, France}
    
    \thanks{25th International Conference on Pattern Recognition (ICPR 2020), p2529–p2535, Milan, Italy, Jan 10-15, 2021}
     
    \IEEEauthorblockA{\IEEEauthorrefmark{2}Mosel LORIA UMR CNRS 7503, Université de Lorraine, 54506, Vandœuvre-lès-Nancy, France \\
    Email: \{weizhu.qian, franck.gechter\}@utbm.fr}
}

\maketitle

\begin{abstract}

Recognizing user location with WiFi fingerprints is a popular approach for accurate indoor positioning problems. In this work, our goal is to interpret WiFi fingerprints into actual user locations. However, WiFi fingerprint data can be very high dimensional in some cases, we need to find a good representation of the input data for the learning task first. Otherwise, using neural networks will suffer from severe overfitting. In this work, we solve this issue by combining the Information Bottleneck method and Variational Inference. Based on these two approaches, we propose a Variational Information Bottleneck model for accurate indoor positioning. The proposed model consists of an encoder structure and a predictor structure. The encoder is to find a good representation in the input data for the learning task. The predictor is to use the latent representation to predict the final output. To enhance the generalization of our model, we also adopt the Dropout technique for each hidden layer of the decoder. We conduct the validation experiments on a real-world dataset. We also compare the proposed model to other existing methods so as to quantify the performances of our method.      

\end{abstract}

\IEEEpeerreviewmaketitle

\section{Introduction}

Localizing smartphone users is a essential technique for many location-based services like navigation and advertisement. Though GPS can provide relatively accurate position, it cannot function well in indoor environment. Thus we need to seek other options to recognize user location. Since the WiFi signal strength is related to distance of the hotspots and the devices, if we have the WiFi fingerprint data labeled with actual user coordinates, then we can interpret WiFi fingerprints into user location information via supervised learning approaches.

%\todo{Can you explain your hypothesis and your constrains}

A lot of research works are focusing on the use of the WiFi received signal strength indicator (RSSI) value data. Among the solutions presented in literature, traditional neural networks are historically among the most widespread. Their limitations are that they can be regarded as deterministic functions in a sense and their loss functions usually are Euclidean distance (for instance, mean squared errors) for regression problems. Conventional neural networks work well in many cases but when the dataset contains too much noisy information, they are not powerful enough to learn the useful information from the dataset. In our case, the user current position is normally only related to a few number of WiFi access points, and the rest of RSSI values in the input vector are in fact not useful. However, in the modelling process, we need to feed all the RSSI values to the model. As a result, the unrelated information in the input data will lead to bad performance when training the neural network.   

Alternatively, some deep probabilistic models can be applied to the aforementioned problems, for instance, Mixture Density Networks (MDNs) \cite{bishop1994mixture}, Bayesian Neural Networks (BNNs) \cite{blundell2015weight} and Variational Autoencoders (VAEs)-based models \cite{qian2019supervised}. These methods are based on the probability theory and Bayesian statistics, by introducing uncertainty to the models to prevent overfitting problems. However, according to the Information Bottleneck theory \cite{tishby2000information}, \cite{alemi2016deep} these models do not consider what the useful information in the dataset are for the learning tasks. Thus, the aforementioned deep probabilistic models can solve our problems in a sense but they may not be the optimal solutions.

In this work, we propose a novel model to calculate the accurate user location by using the related WiFi fingerprints. We treat this problem as a supervised regression problem. It means that we use the WiFi RSSI value data as the input and the actual user location (latitudes and longitudes) as the output. However, there are some difficulties to achieve this goal. First, to provide good quality of network connections, modern building are normally equipped with abundant WiFi access points (WAPs). Therefore, when we use the WiFi received signal strength indicator (RSSI) value data as the modeling input, which usually are very high dimensional. Meanwhile, due to the signal-fading and multi-path \cite{hoang2019recurrent} effects, the RSSI values can be very noisy. These two properties result in severe overfitting when we use conventional neural network-based models.  

For this reason, in contrast with the existing methods, based on the Information Bottleneck method and Variational Inference, we propose a Variational Information Bottleneck model in this work. This model consists of two sub-models, one is the encoder model, the other is the predictor model which is used to predict the target values. According to the Information Bottleneck theory \cite{tishby2000information}, the encoder in our model is used to find a good latent representation of the input data for the related learning task so that the nuisance information in the original input will be token out. Afterwards, the predictor utilizes the latent representation as its input, instead of the original input, to predict the target values. Our model is an end-to-end deep learning model and scalable to large scale datasets which makes it easy to train.

The reminder of the paper is organized as follows. Section~\ref{sec: Related} surveys the related research work. In Section~\ref{sec: Method} introduce the proposed model. Section~\ref{sec: Experiments} demonstrates the validation experiments and the results and gives a detailed discussion. The conclusions and the possible future work are in Section~\ref{sec: Conclusions}.

\section{Related Work}
 \label{sec: Related}

In previous research, both conventional machine learning and deep learning methods are widely explored for WiFi fingerprint based user location recognition problems. Many previous works treat this problem as classification or clustering tasks, which means to identify the buildings and/or floors. Some researchers used conventional machine learning methods, for instance, i.g., Decision Trees, K-nearest neighbors, Naive Bayes, Neural Networks, K-means and the affinity clustering algorithm \cite{bozkurt2015comparative}, \cite{cramariuc2016clustering}, \cite{ferris2007wifi}, \cite{hahnel2006gaussian}, \cite{yiu2015gaussian}, \cite{zhong2019wifi}. In addition, since RSSI values are high dimensional sometimes, some researchers used deep learning techniques like Autoencoders \cite{hinton1994autoencoders} to reduce the input dimension before preceding the main learning tasks \cite{nowicki2017low}, \cite{song2019novel}, \cite{kim2018scalable}.

For learning the accurate user position information, i.e., calculating the real coordinates of the users, Gaussian Processes (GPs) can be one of the options \cite{hahnel2006gaussian}, \cite{ferris2007wifi}, \cite{yiu2015gaussian}. But GPs are extremely computationally expensive when it comes to datasets with large scales because they need to compute the covariances between each data points. To circumvent this issue, one can resort to deep learning approaches. \cite{ibrahim2018cnn}, \cite{song2019novel} and \cite{hoang2019recurrent} used Convolutional Neural Networks (CNNs) and Recurrent Neural Networks (RNNs). However, since deterministic neural networks can cause serve overfitting issue, their methods calculate the use coordinates indirectly. In our study, we find that some deep probabilistic models can be better solutions. For example, Mixture Density Networks \cite{bishop1994mixture} use a set of mixed Gaussian distributions at the output layers to compute the final output and use the negative log-likelihood as the loss function. The disadvantage of such method is that, as a maximum likelihood estimation (MLE) method and it fails to consider the prior of the model parameters so it is prone to be overfitting. Though Bayesian Neural Networks (BNNs) \cite{blundell2015weight} are maximum a posteriori (MAP) methods, they do not extract the noisy information in the input data, so their performance is not good as expected either. In our previous research, we take advantage of Variational Autoencoders (VAEs) \cite{kingma2013auto}, \cite{rezende2014stochastic} to develop the VAE-based semi-supervised learning models \cite{qian2019supervised}. However, these models all neglect the effects of the nuisance information in the dataset. The nuisance information is redundant for the learning tasks and will damage the modeling performance. To circumvent the effect of the nuisance information, one can resort to the Deep Variational Information Bottleneck (DVIB) model \cite{alemi2016deep}. DVIB is a model based on Variational Inference and the Information Bottleneck method. It aims at learning a good latent representation of the input data for the downstream learning tasks.

In this work, we want to apply the Information Bottleneck method to WiFi fingerprint-based location recognition problem in order to reduce the nuisance information damaging the modeling performance. Inspired by VAEs, $\beta$-VAEs \cite{higgins2017beta}, \cite{burgess2018understanding} and DVIB, we devise a Variational Information Bottleneck model to interpret the user WiFi RSSI values into the actual user location information. This model is solved via Monte Carlo sampling and Variational Inference.

\section{Method}
\label{sec: Method}

\subsection{Preliminaries}
In our model, the input is the WiFi RSSI values $x$, the output is the user's coordinates $y$. To make the model more robust to noise, we use a set of probabilistic distributions such as $p(z|x)$ and $p(y|z)$ to describe the relationship between the variables instead of deterministic functions as in conventional neural networks. Furthermore, in order to let our model work theoretically, we need to make some assumptions first:
\begin{itemize}
    \item \textbf{Assumption $1$}: assume that there exists a latent distribution of $z$. Let's say that $x$, $z$, $y$ belong to the same information Markov chain: $x \rightarrow z \rightarrow y$.
    \item \textbf{Assumption $2$}: assume that $x$ is solely sufficient enough to learn $z$, which leads to $p(z|x,y) = p(z|x)$.
    \item \textbf{Assumption $3$}: assume that $z$ is solely sufficient enough to learn $y$, which leads to $p(y|x,z) = p(y|z)$.
\end{itemize}

We make the above assumptions based on the idea that the values of both the WiFi RSSIs and GPS coordinates are related to the user's real physical position. Hence either the WiFi RSSI values or the GPS coordinates contains the sufficient information of the real user physical position (which we use a the latent variable $z$ to represent). This suggests that we can use $x$ to compute $p(z|x)$ (encoding step) and then use $y$ to compute $p(y|z)$ (predicting step). The above assumptions will facilitate the derivation of our model.   

\subsection{Model}

In a maximum a posteriori (MAP) modeling setting, the parameters of the model are related to not only the dataset but also the prior of the parameters:
\begin{align}
\label{Eq: MAP}
p(\lambda|D) \propto p(D|\lambda) q(\lambda) 
\end{align}
where, $D$ is the dataset, $\lambda$ is the model parameters, $p(\lambda|D)$ is the posterior, $p(D|\lambda)$ is the likelihood and $q(\lambda)$ is the prior. Applying such a setting to our problem, the prior of the latent representation $z$, $q(z)$ and the posterior $p(z|x)$ can both be represented by Gaussian distributions. Through using Variational Inference, $p(z|x)$ can be calculated via a neural network.  

In Variational Autoencoders, one assumes that there is a latent distribution of $z$ which can be used to reconstruct the original input $x$. Hence the information Markov chain of VAEs is $x \rightarrow z \rightarrow x'$, where $x'$ is the reconstructed input. Accordingly, the loss function can be written as :
\begin{align}
\label{Eq: OP_VAE}
\mathcal{L}(D,\theta,\phi) & ~ = \mathop{{}\mathbb{E}}{_{z \sim p_{\phi}(z|x)}}[p_{\theta}(x|z)]  \nonumber \\ 
& \qquad -  D_{KL}[p_{\phi}(z|x)|| q(z)]  
\end{align}

where $D_{KL}$ represents the Kullback–Leibler (KL) divergence, which is to measure the closeness between the posterior and the prior, $\phi$ is the parameters of the encoder network, $\theta$ is the parameters of the decoder network, $q(z)$ is an uninformative prior of $z$, here we can use a standard Normal distribution $\mathcal{N}(0,\mathbb{I})$.

Furthermore, according to the Information Bottleneck principle \cite{tishby2000information}, let $x$ be the input, $y$ be the learning target and $z$ be the representation, then we can have the following optimization objective:
\begin{align}
\label{Eq: IB}
\text{max}~I(z;y) ~\text{s.t.}~ I(z;x) \leq I_c
\end{align}

where $I$ is the mutual information, $I_c$ is the information constraint.

Or equivalently, if we apply the Karush-Kuhn-Tucker (KKT) conditions to Eq.~(\ref{Eq: IB}), then we will have the following Lagrangian:
\begin{align}
\label{Eq: IB_L}
\mathcal{L}_{IB} = I(z;y)-\beta I(z;x)
\end{align}

where $\beta$ is a Lagrangian multiplier.

$\beta$-VAEs leverage Eq.~(\ref{Eq: IB_L}) to formulate a constrained variational framework, so that Eq.~(\ref{Eq: OP_VAE}) becomes:         
\begin{align}
\label{Eq: OP_B_VAE}
\mathcal{L}(D,w,\phi) & ~ = \mathop{{}\mathbb{E}}{_{z \sim p_{\phi}(z|x)}}[p_{w}(x|z)]  \nonumber \\ 
& \qquad -  \beta D_{KL}[p_{\phi}(z|x)|| q(z)]  
\end{align}

Since our learning task is supervised, as opposed to VAEs and $\beta$-VAEs, we have the information Markov chain: \\ $x \rightarrow z \rightarrow y$.
As opposed to $\beta$-VAEs \cite{higgins2017beta}, \cite{burgess2018understanding}, based on Eq.~(\ref{Eq: IB}) and the assumptions we have made, we know that the latent variable $z$ can be represented by $x$ alone ($p(z|x,y)= p(z|x)$) and the output $y$ can depend on $z$ alone ($p(y|x,z)= p(y|z)$). For this reason, we can replace the term $p(x|z)$ in Eq.~(\ref{Eq: OP_VAE}) with $p(y|z)$. As a result, now we have this following optimization objective for our model:  
\begin{align}
\label{Eq: OP_VIB}
\underset{\theta,~\phi}{\mathrm{argmax}} & ~ \mathop{{}\mathbb{E}}{_{D}}[ \mathop{{}\mathbb{E}}{_{p_{\phi}(z|x)}}[\log p_{w}(y|z)]] \nonumber \\
& ~ s.t.~ D_{KL}[p_{\phi}(z|x), q(z)] \leq \epsilon 
\end{align}

where $D=\{x,y\}$ is the dataset, $\theta$ is the parameters of the predictor network, $\epsilon$ is a positive constant with small value.

Further, based on Eq.~(\ref{Eq: IB_L}), the Lagrangian form of Eq.~(\ref{Eq: OP_VIB}) can be written as:  
\begin{align}
\label{Eq: OP_VIB_L}
\mathcal{L}(D,w,\phi) & ~ = \mathop{{}\mathbb{E}}{_{z \sim p_{\phi}(z|x)}}[p_{w}(y|z)]  \nonumber \\ 
& \qquad- \beta D_{KL}[p_{\phi}(z|x)|| q(z)]  
\end{align}

\begin{figure}[t]%{1.\textwidth}
\centering
\includegraphics[width= \linewidth]{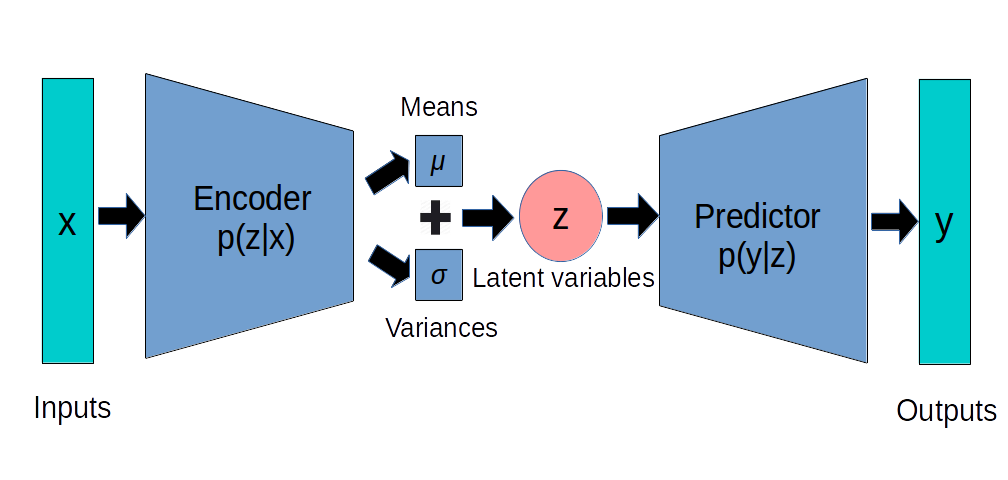}
\caption{Comparisons.}\label{Fig: DVIB}    
\end{figure}

Finally,  Eq.~(\ref{Eq: OP_VIB_L}) is the loss function of the proposed model. In contrast with VAEs and $\beta$-VAEs, which are unsupervised learning models, our model is an end-to-end supervised learning model. As shown in Fig~\ref{Fig: DVIB}, $p_{\phi}(z|x)$ represents the encoder neural network and $p_{w}(y|z)$ represents the predictor neural network. 

\subsection{Model Solver}
To solve the Eq.~(\ref{Eq: OP_VIB_L}), we need to adopt some special techniques. First, for computing the term $D_{KL}[p{\phi}(z|x)||q(z)]$, we can use the reparameterization trick proposed in \cite{kingma2013auto}, in which the random distribution of $z$ is decomposed as the combination of the mean $\mu$ and the variance $\sigma$:    
\begin{align}
\label{Eq: Repara}
z = \mu_z+ \sigma_z \odot \epsilon_z
\end{align}

where, $\mu_z$ and $\sigma_z$ can be calculated via the neural networks respectively, and $\epsilon_z$ can be sampled from a standard diagonal Normal distribution.

Afterwards we need to calculate the term $\mathop{{}\mathbb{E}}{_{z \sim p_{\phi}(z|x)}}[p_{w}(y|z)]$. This term cannot be solved directly but we can use Monte Carlo method to compute it. 

If we adopt Monte Carlo sampling, then Eq.~(\ref{Eq: OP_VIB_L}) becomes:
\begin{align}
\label{Eq: Loss_data_re}
\mathcal{L}(D,w,\phi) = \frac{1}{N}\sum_{n=1}^{N}\mathop{{}\mathbb{E}_{ \epsilon_z \sim p(\epsilon_z)}}[p_{w}(y_n|f_\phi(x_n,\epsilon_z))]  \nonumber \\ - \beta D_{KL}[p_{\phi}(z|x_n)||q(z)]  
\end{align}

where $N$ denotes the total instance number. $f_\phi(x)$ is the same deterministic neural network used in the encoder to calculate the parameters of the distribution $p(z|x)$:
\begin{align}
\label{Eq: F_determistic}
f_\phi(x) = \mu_z(x) + \sigma_z(x) \odot \epsilon_z
\end{align}

Last but not least, $\beta$ is a hyperparameter which is used to balance the encoding term and the predicting term so that it needs to be chosen carefully. 

\subsection{Computing Output}

In VAEs and $\beta$-VAEs, one can obtain new samples from an uninformative standard Gaussian first then use them as the input of the decoder. Whereas since our model is a supervised model, once the model is trained, we use the sample from the conditional distribution, i.e., $p(z|x)$, to feed the predictor network to compute the final output, which is the same as the training procedure.    

\begin{algorithm} 
\caption{Algorithm}
\label{Alg: Algorithm}
\begin{algorithmic}[1]
\Require{$X$ (inputs), $Y$ (targets)} 
\Ensure{$Y'$ (predictions)}
\Statex

\While{ $\text{epoch} \leq \text{max epoch}$}
    
    \State{$\mu_z$, $\sigma_z$ $\gets$ $E_\phi(X)$}  \Comment{$E_\phi(X)$: Encoder network}
    \State{$z \sim \mathcal{N}(\mu_z, \sigma_z)$}  \Comment{Sample latent codes}
    \State{$Y'$ $\gets$ $F_y(z;w)$}    \Comment{$F_y$: Predictor network}
    \State{minimize loss function $\mathcal{L}(D,w,\phi)$}   \Comment{Eq.~(\ref{Eq: OP_VIB_L})}
    
\EndWhile\\
\Statex
\Return{$Y'$}

\end{algorithmic}
\end{algorithm}

The overall algorithm is summarized in Algorithm~\ref{Alg: Algorithm}.

\section{Experimental Results}
\label{sec: Experiments}

\subsection{Dataset Description}
For the validation, we use the UJIindoor dataset \cite{torres2014ujiindoorloc} whose input dimension is $520$ and each dimension represents a WAP. The RSSI values range from $-110$ dB to $0$ dB when the WAPs are detected, otherwise the RSSI values are set to be $100$. Also each RSSI vector corresponds to a pair of latitude and longitude as the geo-location label. In our experiments, we use scaled GPS coordinates values for computational convenience. The total instance number is about $20000$. For Experiment 1 and Experiment 2, we use $80\%$ of the dataset for training and the rest $20\%$ as the test dataset. In Experiment 3, the training data number will vary.      

\subsection{Model Structure}
\begin{table*}%[h]
\centering
\caption{Model Implementation Details}
\label{Tab: DVIB_Imp}
\begin{tabular}{|c|c|c|c|}
\hline
Sub-network & Layer & Parameter & Activation Function\\
\hline
Encoder & hidden layer & neuron number: 512; latent dimension: 5  & ReLU\\
Predictor & hidden layer & neuron number: 512: dropout rate: 0.3  & ReLU \\
Predictor & hidden layer & neuron number: 512: dropout rate: 0.3  & ReLU \\
Predictor & hidden layer & neuron number: 512: dropout rate: 0.3  & ReLU \\
\hline

\multicolumn{4}{|c|}{Optimizer: Adam; learning rate: 1e-3} \\

\hline
\end{tabular}
\end{table*}

Table~\ref{Tab: DVIB_Imp} demonstrates the implementation details of our model. The encoder neural network includes of one hidden layer, and the dimension of the latent codes is set to be $5$. In practice, we find that the latent dimension of $5$ is in line with the Minimal Description Length principle \cite{hinton1994autoencoders} for our task. The predictor is composed of three hidden layers. Each hidden layer has $512$ units. Especially, in order to improve modeling generalization on test data, we can increase the model uncertainty. Hence we apply the Dropout technique \cite{dahl2013improving} to the hidden layers of the predictor. The optimizer for the model is Adam \cite{kingma2014adam} and the learning rate is $1\text{e-}3$.

\subsection{Experiment 1}

In the loss function of the proposed model, the constant $\beta$ is related to the constraint for the optimization, which is to balance the encoding error term $\mathop{{}\mathbb{E}}{_{z \sim p_{\phi}(z|x)}}[p_{w}(y|z)]$ and the prediction error term $D_{KL}[p_{\phi}(z|x)||q(z)]$. A larger $\beta$ value means the model tends to be more compressive for the input and less expressive for the output, and vice versa. Therefore, different $\beta$ values can result in different modeling results. 

\begin{figure}[t]%{1.\textwidth}
\centering
\includegraphics[width= \linewidth]{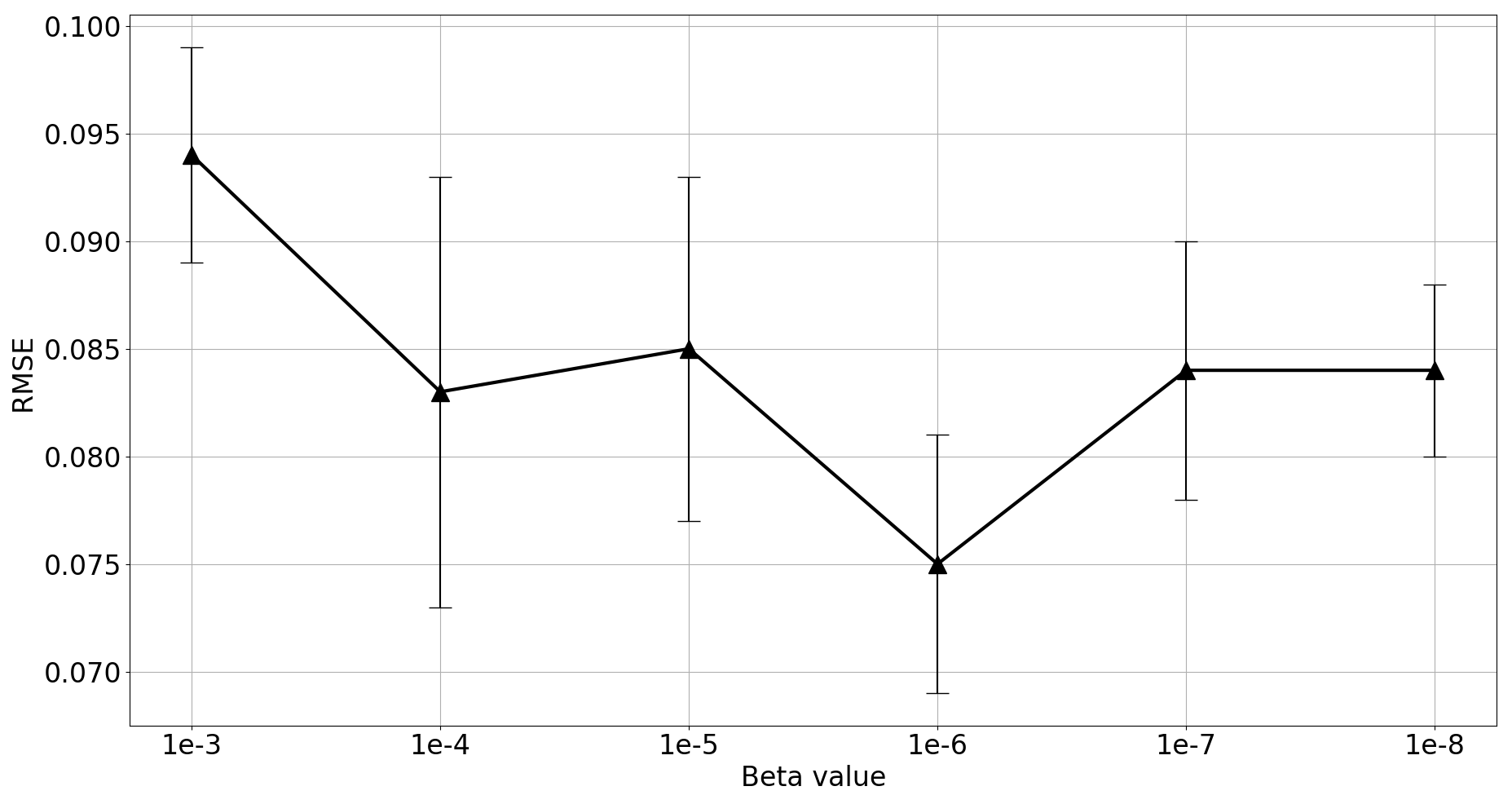}
\caption{Changing the value of $\beta$.}\label{Fig: Beta}    
\end{figure}

To find the optimal $\beta$ values, we test different $\beta$ values, ranging from $1\text{e-}3$ to $1\text{e-}8$, for our model. From the results shown in Fig.~\ref{Fig: Beta}, we can see that, when $\beta$ is $1\text{e-}6$, the proposed model has the best performance. Thus, we will hereafter set $\beta$ to be $1 \text{e-}6$ for the propose model in all following experiments.

\begin{figure}%[!t]
\centering
    \begin{subfigure}[t]{0.4\textwidth}
    \includegraphics[width= \linewidth]{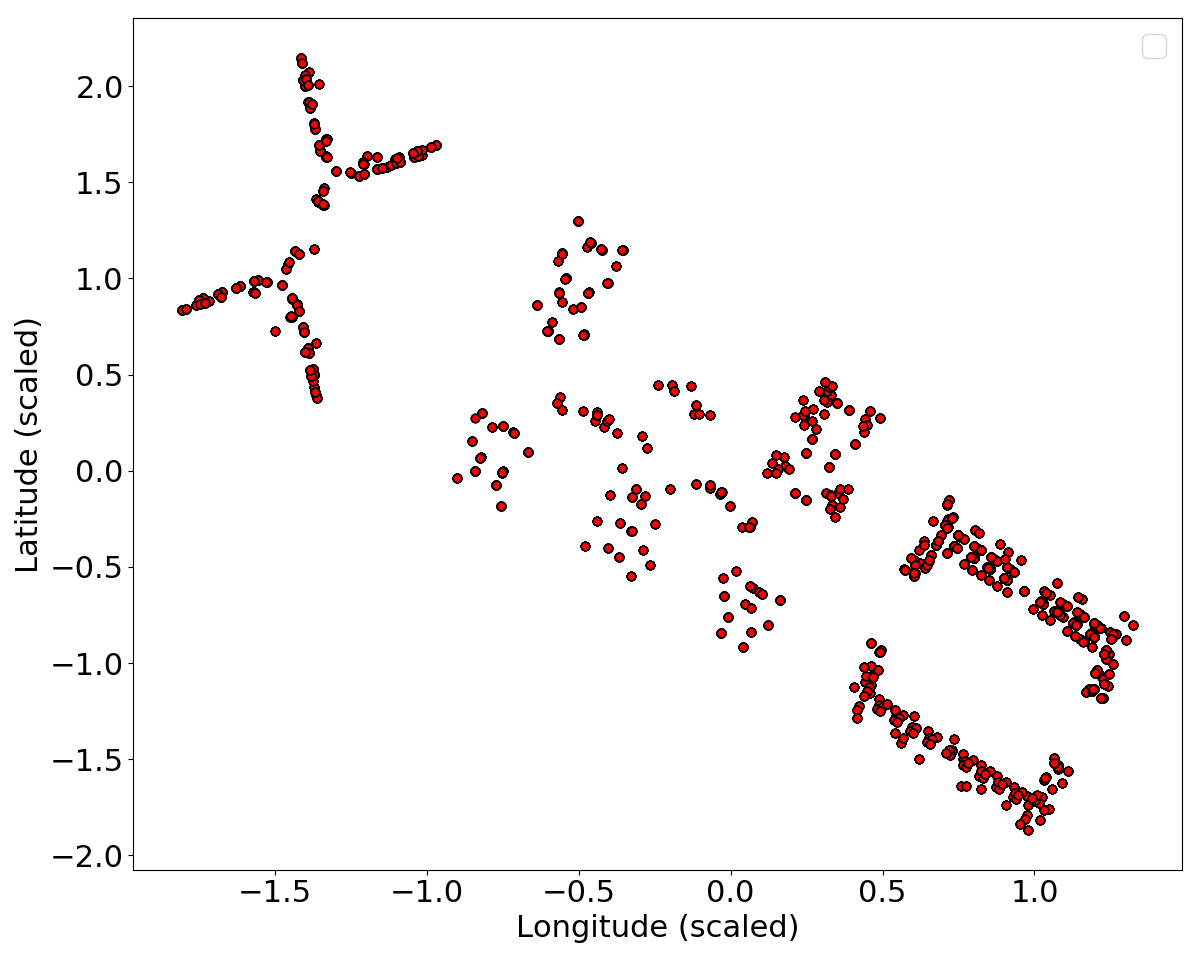}
    \caption{Ground truth.}
    \label{fig: Ground_truth}
    \end{subfigure}

    \begin{subfigure}[!t]{0.4\textwidth}
    \centering
    \includegraphics[width= \linewidth]{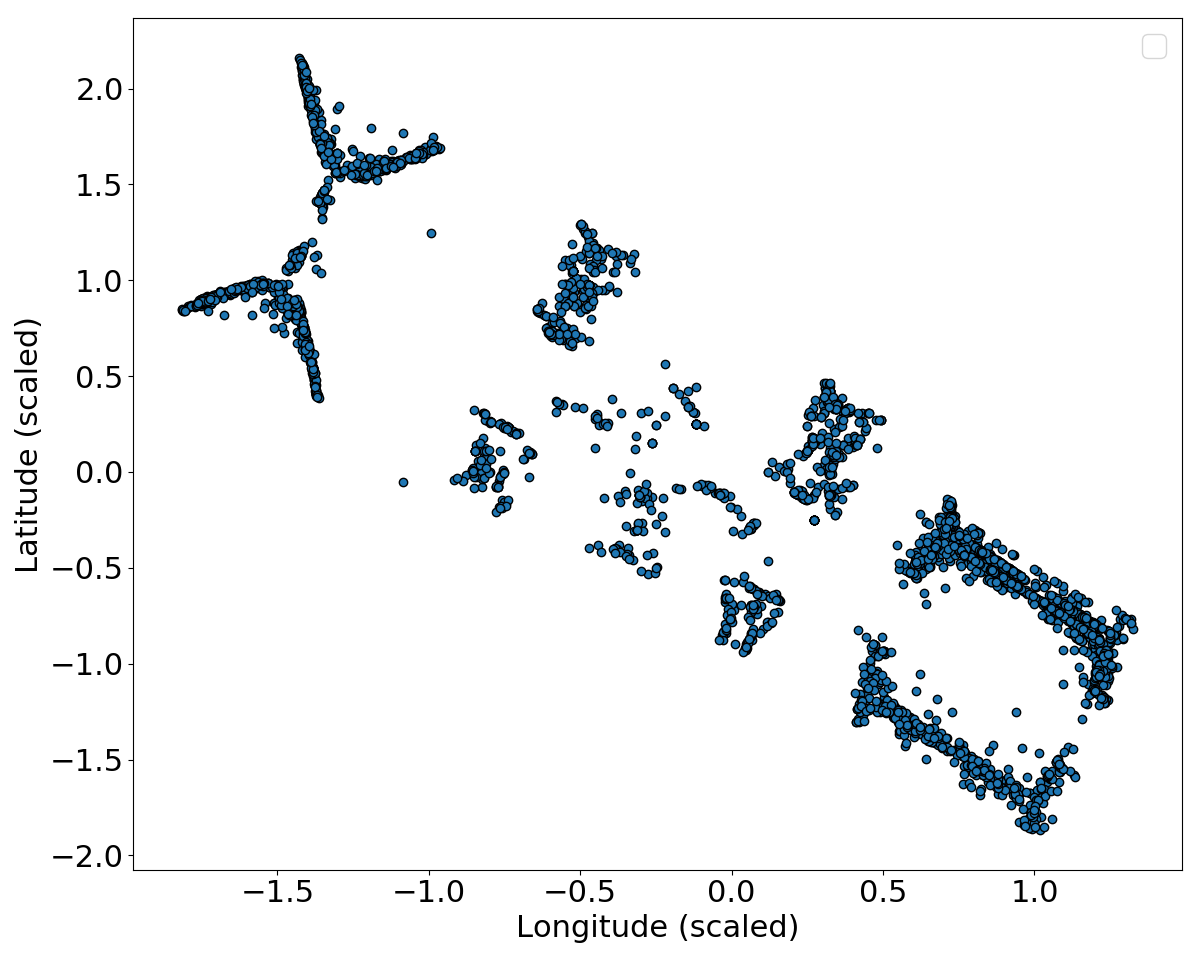}
        \caption{Test result.}
    \label{fig: Result}
    \end{subfigure}
    
\caption{Experimental results.} 
\label{fig: Results}
\end{figure}

\begin{figure}%[!t]
\centering
    \begin{subfigure}[t]{0.4\textwidth}
    \includegraphics[width= \linewidth]{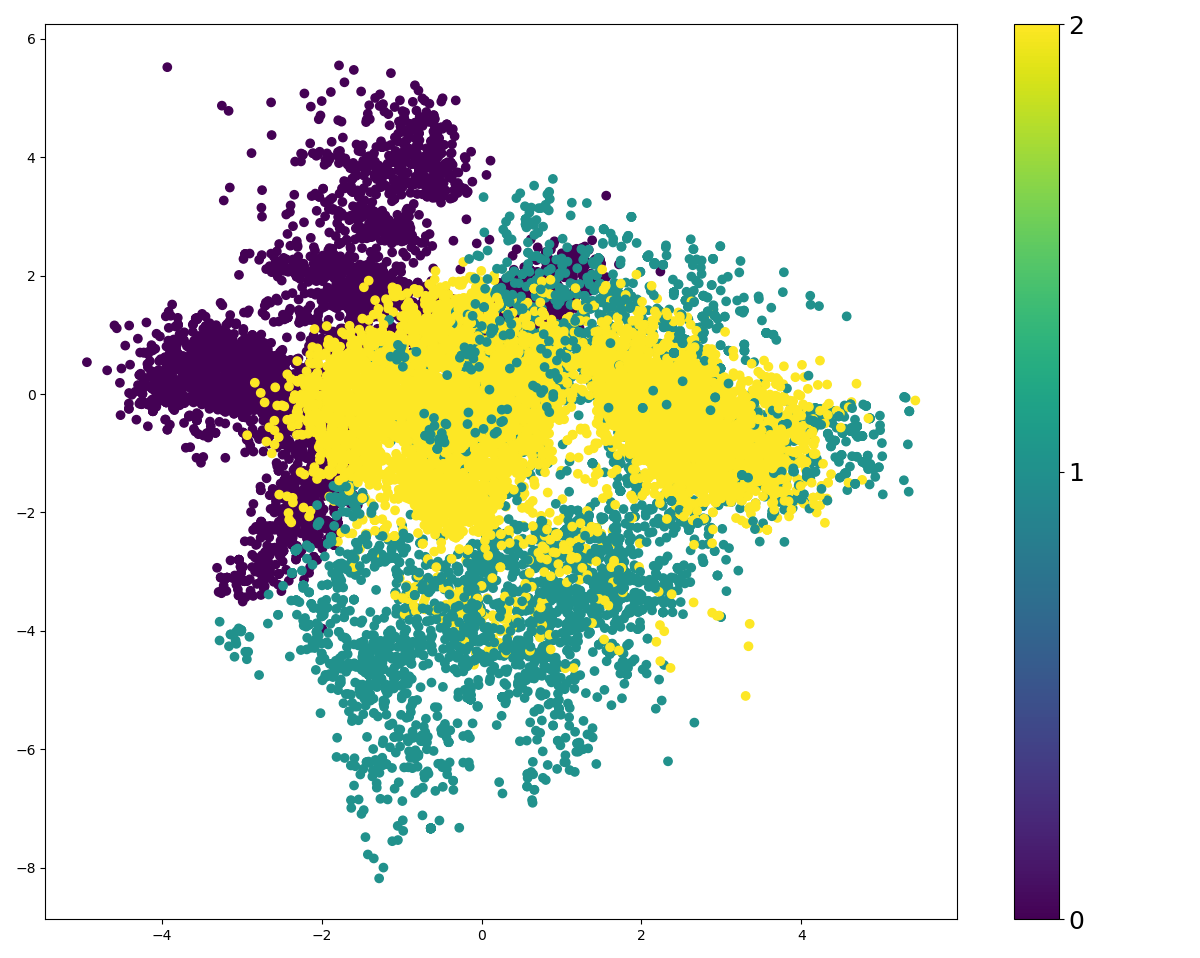}
    \caption{Latent variables labeled with the building IDs.}\label{Fig: Z_B}
    \end{subfigure}

    \begin{subfigure}[!t]{0.4\textwidth}
    \centering
    \includegraphics[width= \linewidth]{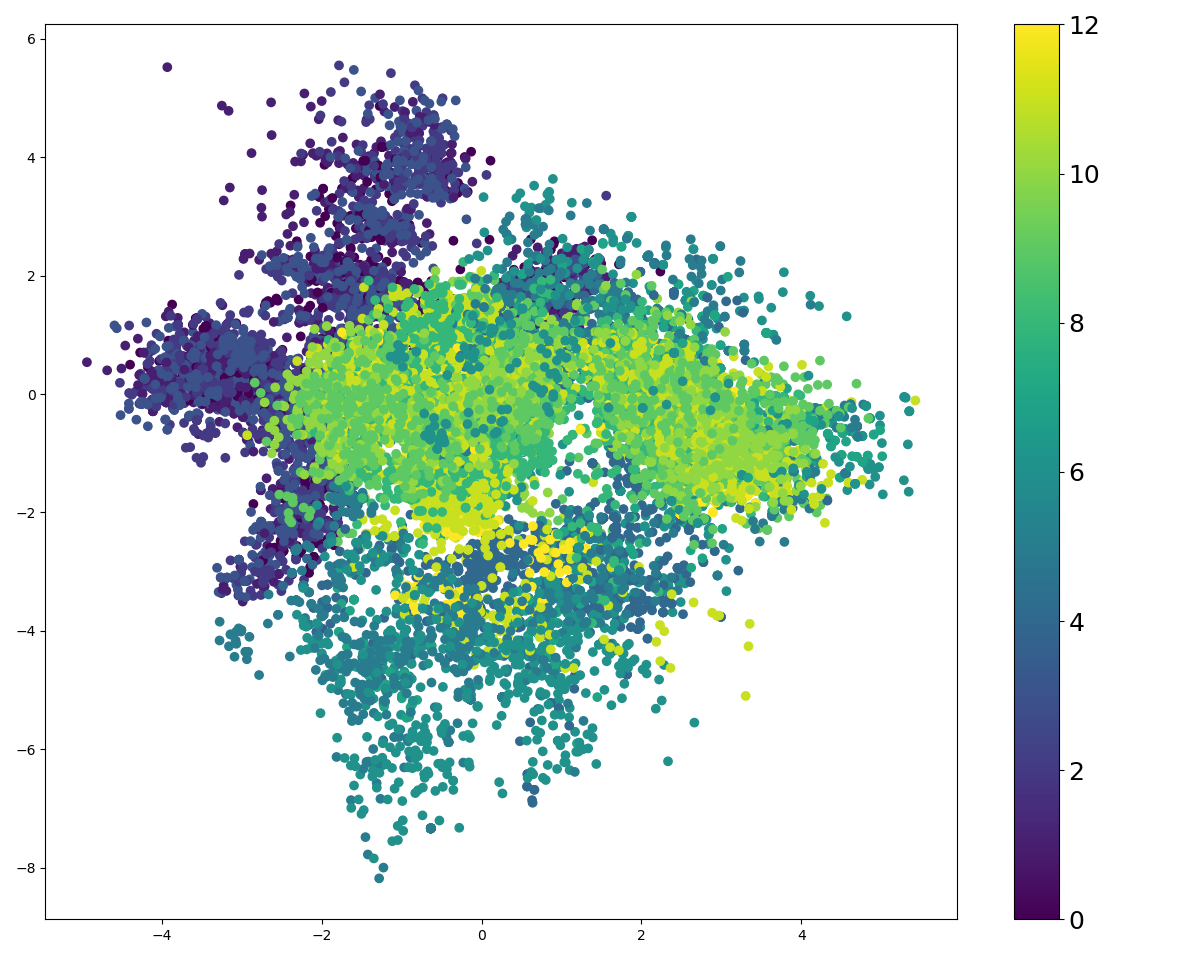}
    \caption{Latent variables labeled with the floor IDs.}\label{Fig: Z_F}
    \end{subfigure}
    
\caption{Latent variables with dimension of $5$, here shows the 2D projection.}
\label{Fig: Latent}    
\end{figure}

Fig.~\ref{fig: Results} shows the ground truth and the test modeling result of our model. It can be seen the proposed model can calculate the user location coordinates accurately using the relevant WiFi fingerprints. In addition, Fig.~\ref{Fig: Latent} demonstrates how the latent distribution is related to the building IDs and floor IDs.

\subsection{Experiment 2}
\begin{table*}
\centering
\caption{Comparison Results}
\label{Tab: Comparisons}
\begin{tabular}{|c|c|c|c|c|c|c|c|}
\hline
Method & k-NN & GP & MDN-2 & MDN-5 & BNN & Semi-VAE & Proposed\\
\hline
RMSE & $0.092\pm2\text{e-}3$ & $0.252\pm3\text{e-}3$ & $0.099\pm3\text{e-}4$  & $0.103\pm3\text{e-}3$& $1.033\pm4\text{e-}3$ & $0.088\pm1\text{e-}3$ & $0.075\pm 6\text{e-}3$\\
\hline
\end{tabular}
\end{table*}

To show the advantages of our method, we run other methods proposed in the literature on the UJIindoor dataset. K-NN is used as the baseline model. The MDN-2 model refers to the Mixture Density Network model with $2$ Gaussian distributions at the output layers. Similarly, the MDN-5 model is a MDN model with $5$ Gaussian distributions at the output layers. The Semi-VAE model is a semi-supervised variational autoencoder (VAE) model, which will be explained later. The overall results are shown in Table~\ref{Tab: Comparisons}. We use the root mean squared error (RMSE) as the evaluation metrics.          

From the results, we can see that the proposed model has the best modeling performance. Also in practice we find that compared to our model, the Gaussian Process model suffers from heavy computation load and the MDN models are not very stable during the learning process.

\subsection{Experiment 3}
From our previous assumptions, as an alternative approach, we can also formulate a semi-supervised learning approach, the semi-VAE model. The learning procedure can be described briefly as follow. If we learn a VAE model via unsupervised model at first, then we will have $p(z|x)$ and $p(x|z)$. After that we can do a supervised learning procedure, by sampling $z$ from $p(z|x)$ to compute $p(y|z)$. Especially, in the semi-VAE model, the model uses both the labeled and unlabeled data for unsupervised learning and then uses the labeled data for supervised learning. While, in our proposed model, we only use the labeled data for supervised learning.     

\begin{figure}[t]%{1.\textwidth}
\centering
\includegraphics[width= \linewidth]{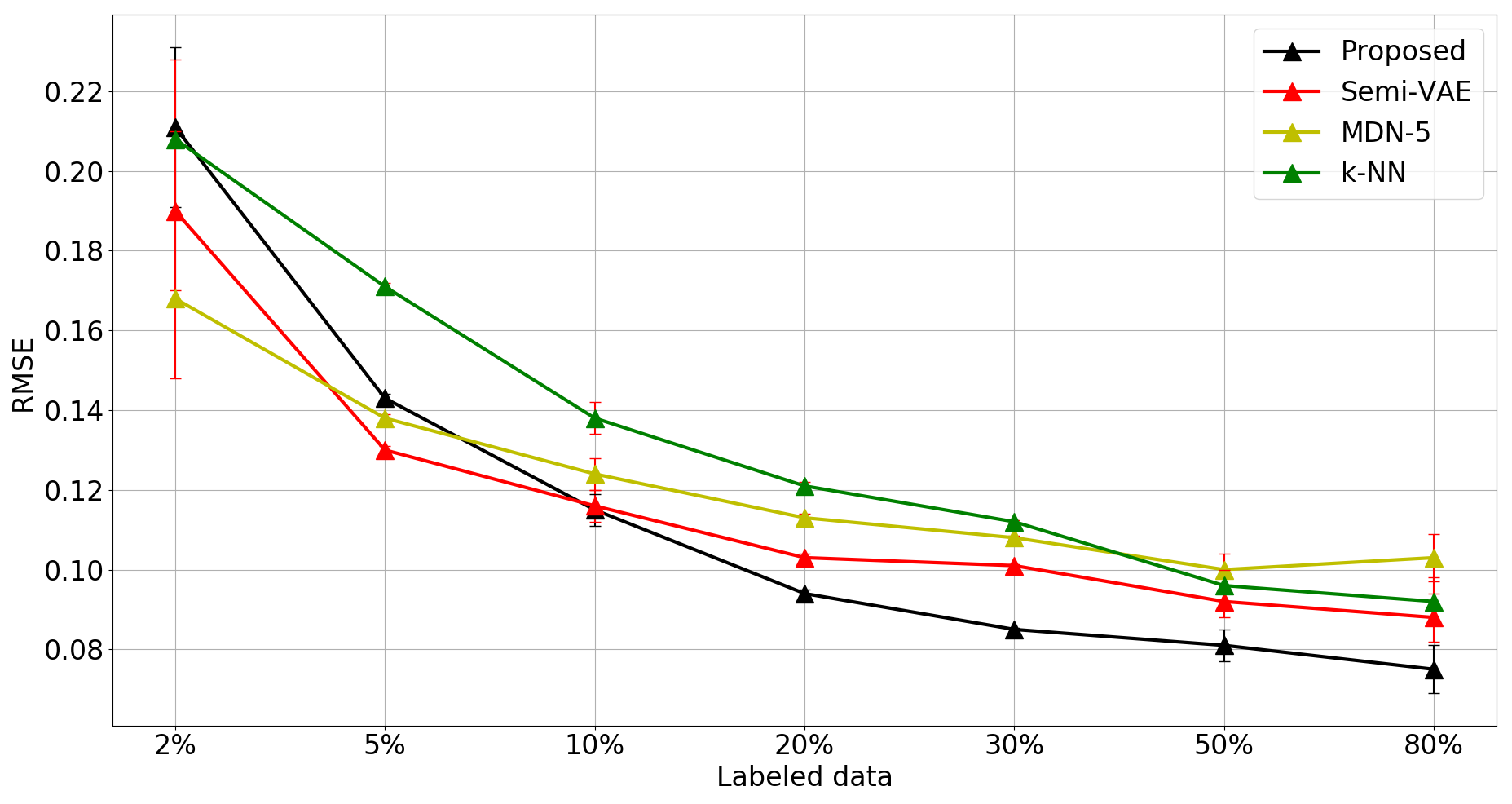}
\caption{Results on different portions of labeled data.}\label{Fig: Semi}    
\end{figure}

To compare with the semi-supervised learning approach, the semi-VAE model, we run our model and other models on different portions of labeled data. As shown in Fig.~\ref{Fig: Semi}, we can see that once the labeled data are more than $10\%$ of the total training data, our method surprisingly has the best performance among all the methods.

\subsection{Discussion}
\label{sec: Discussion}
Why the proposed method can outperform other deep learning methods? First, our problem can be regarded as a regression problem, and especially, the input (RSSI vectors) is relatively high dimensional and the target (GPS coordinates) is rather low dimensional. Thus, it causes the issue that the input has redundant information for the learning tasks. If we use a conventional Neural Network to solve this problem directly, the results will not be satisfying at all. Mixture Density Networks and Bayesian Neural Networks handle this problem by introducing uncertainty into the models. The difference is that MDNs are MLE method while BNNs are MAP method. Surprisingly, The BNN has worse performance than the MDNs on our tasks because the uncertainty of BNNs does not depend on the input data. Variational Autoencoders are originally designed as generative approaches to obtain new sample data. For our problem, we can use a VAE to learn the latent representation of the input data first. Then this model can be trivially extended to be a semi-supervised model by using the pre-learned representation to obtain the final output. However, in our study, we find that leveraging the Information Bottleneck method to this problem is a better option than the semi-VAE model. It is because that, with the Information Bottleneck method, we can treat the original task as a constrained optimization problem. The optimization objective is the learning tasks and the constraint is the data representation. That's to say the Variational Information Bottleneck Model is to directly find the optimal representation for the learning tasks, whereas the semi-VAE model is to find the representation to reconstruct the original inputs.

\section{Conclusions}
\label{sec: Conclusions}

Interpreting WiFi fingerprints into real user location via neural networks is a tricky problem. In this work, we combined the Information Bottleneck theory with Variational Inference to propose a novel deep learning model for WiFi fingerprint-based user location recognition. The proposed model consists of two neural networks, an encoder and a predictor. According to the Information Bottleneck theory, the encoder neural network is to find an optimal representation of the input data and mitigate the negative effect of the nuisance information for the learning tasks. The predictor neural network is to use the latent representation to compute the final output. The main advantages of the proposed model are that it is scalable to large scale dataset, computationally stable and robust to noisy information. To evaluate our model, we run our model and other previous models on the real-world WiFi fingerprint dataset and the finally results verify the effectiveness and show the advantages of our method compared to the existing approaches. For the future research, we plan to explore other methods in information theory and Variational Inference to improve the performance of our models or develop other applications.

%\vfill\pagebreak

\section*{Acknowledgment}
The authors would like to thank the China Scholarship Council for the financial support.

\bibliographystyle{plain}
\bibliography{refs}

\end{document}